\documentclass{article}

\PassOptionsToPackage{numbers,square, compress}{natbib}

\usepackage[final]{neurips_2022}

\usepackage{bibspacing}

\usepackage[utf8]{inputenc} %
\usepackage[T1]{fontenc}    %
\usepackage{hyperref}       %
\usepackage{url}            %
\usepackage{booktabs}       %
\usepackage{amsfonts}       %
\usepackage{nicefrac}       %
\usepackage{microtype}      %
\usepackage{xcolor}         %

\usepackage{float}
\usepackage{graphicx}
\usepackage{comment}
\usepackage{amsmath,amssymb} %
\usepackage{color}
\usepackage{xspace}
\usepackage{textcomp}
\usepackage{mathtools}
\usepackage{bbm}
\usepackage{makecell, verbatim, sidecap}
\usepackage{subfig}
\usepackage{multirow}
\usepackage{nicefrac}
\usepackage{gensymb}

\hypersetup{
    colorlinks=true,
    breaklinks=true,
    urlcolor=blue,
    linkcolor=blue,
    bookmarksopen=false,
    filecolor=black,
    citecolor=blue,
    linkbordercolor=blue
}

\def\eg{{\it e.g.}\xspace}
\def\ie{{\it i.e.}\xspace}

\def\R{{\mathbb R}}

\def\atan2{{\rm atan2}}
\def\log{{\rm log}}

\def\softmax{{\rm softmax}}

\def\Ours{FCOS-LiDAR\xspace}

\title{Fully Convolutional One-Stage 3D Object Detection on LiDAR Range Images}
\author{
\textbf{Zhi Tian}$^1$,
~~
\textbf{Xiangxiang Chu}$^1$,
~~
\textbf{Xiaoming Wang}$^2$\thanks{Work done as an intern at Meituan Inc.},
~~
\textbf{Xiaolin Wei}$^1$,
~~
\textbf{Chunhua Shen}$^3$\thanks{Corresponding author.}
\\ [0.25cm]
$^1$ Meituan Inc. ~ ~ ~ ~   $^2$ 
Northwestern Polytechnical University
 ~ ~ ~ ~   $^3$ Zhejiang University
\\[0.1cm]
{$^1$ \tt\small \{tianzhi02,chuxiangxiang,weixiaolin02\}@meituan.com}\\
{$^2$ \tt\small chunhua@me.com}
~~
{$^3$ \tt\small xiaomingwang80@163.com}
}

\begin{document}

\maketitle

\begin{abstract}
We present a simple yet effective fully convolutional one-stage 3D object detector for LiDAR point clouds of autonomous driving scenes, termed \Ours. Unlike the dominant methods that use the bird-eye view (BEV), our proposed detector detects objects from the range view (RV, \textit{a.k.a.}\ range image) of the LiDAR points. Due to the range view's compactness and compatibility with the LiDAR sensors' sampling process on self-driving cars, the range view-based object detector can be realized by solely exploiting the vanilla 2D convolutions, departing from the BEV-based methods which often involve complicated voxelization operations and sparse convolutions.
  
For the first time, we show that an RV-based 3D detector with standard 2D convolutions alone can achieve comparable performance to state-of-the-art BEV-based detectors while being significantly faster and simpler. More importantly, almost all previous range view-based detectors only focus on single-frame point clouds, since it is challenging to fuse multi-frame point clouds into a single range view. In this work, we tackle this challenging issue with a novel range view projection mechanism, and for the first time demonstrate the benefits of fusing multi-frame point clouds for a range-view based detector. Extensive experiments on nuScenes show the superiority of our proposed method and we believe that our work can be strong evidence that an RV-based 3D detector can compare favourably with the current mainstream BEV-based detectors.
\end{abstract}

\section{Introduction}
With the rise of autonomous driving, 3D object detection from the LiDAR point cloud has been recently drawing increasing attention. Similar to the 2D image object detection~\cite{ren2015faster, tian2019fcos, liu2016ssd, redmon2017yolo9000, redmon2016you}, 3D object detection requires the model to predict the (3D) locations of the objects of interest and the associated properties (\eg, categories, sizes, heading, and the state of motion). In spite of the unprecedented success that the computer vision community has attained on the 2D image object detection, it is still intractable to transfer the success to the 3D object detection task.

Most %
previous 3D object detection methods consider that the point cloud is amorphous and consists of a set of unordered points. Thus, this task is considered significantly different from its 2D detection counterpart, which works on 
structured RGB images. Moreover, the cornerstones of %
modern computer vision---CNNs or CNN-like vision transformers (\eg, Swin Transformers~\cite{liu2021swin}) also assume 
that 
the inputs are well-organized as grids, which poses another difficulty of accurate 3D object detection in point clouds. As a result, in order to adopt these well-developed techniques, almost all %
top-performing point cloud object detection methods first partition the 3D space into \textit{structured} voxels (or pillars)~\cite{zhou2018voxelnet, yin2021center, yan2018second, lang2019pointpillars, li20173d} and then follow a paradigm similar to
that of 
2D object detectors~\cite{ren2015faster, zhou2019objects}. Additionally, given the prior that it is very rare two objects 
being 
stacked along with the elevation axis in autonomous driving scenes, most %
methods only carry out the detection task on the bird-eye view (BEV) of the point cloud, which can reduce the exponentially increasing complexity resulted from the third dimension. However, owing to incompatibility with the LiDAR's sampling process in the autonomous driving scenes, %
BEV-based solutions %
often suffer from 
the following shortcomings. 1) In fact, the points in autonomous driving are regularly sampled in the spherical coordinate system with the origin being the LiDAR sensor. The BEV disregards this regularity, and causes the issue that the voxels far away from the origin have much fewer points than the ones near the origin; and a large number of voxels are even empty. This results in the need for sparse convolutions~\cite{yan2018second, yin2021center, graham2014spatially}, significantly complicating the system, particularly for on-device applications. 2) The points in a voxel have to be sampled or padded so that every voxel has the same number of points. The sampling decimates a large number of points, leading to the loss of information before the model see anything. 3) From the BEV, some objects such as ``pedestrian'' and ``traffic cone'' become very small. Accurately detecting these objects requires a fine-grained voxel size, dramatically increasing the price of computation.

\begin{figure}
  \centering
  \includegraphics[width=\textwidth]{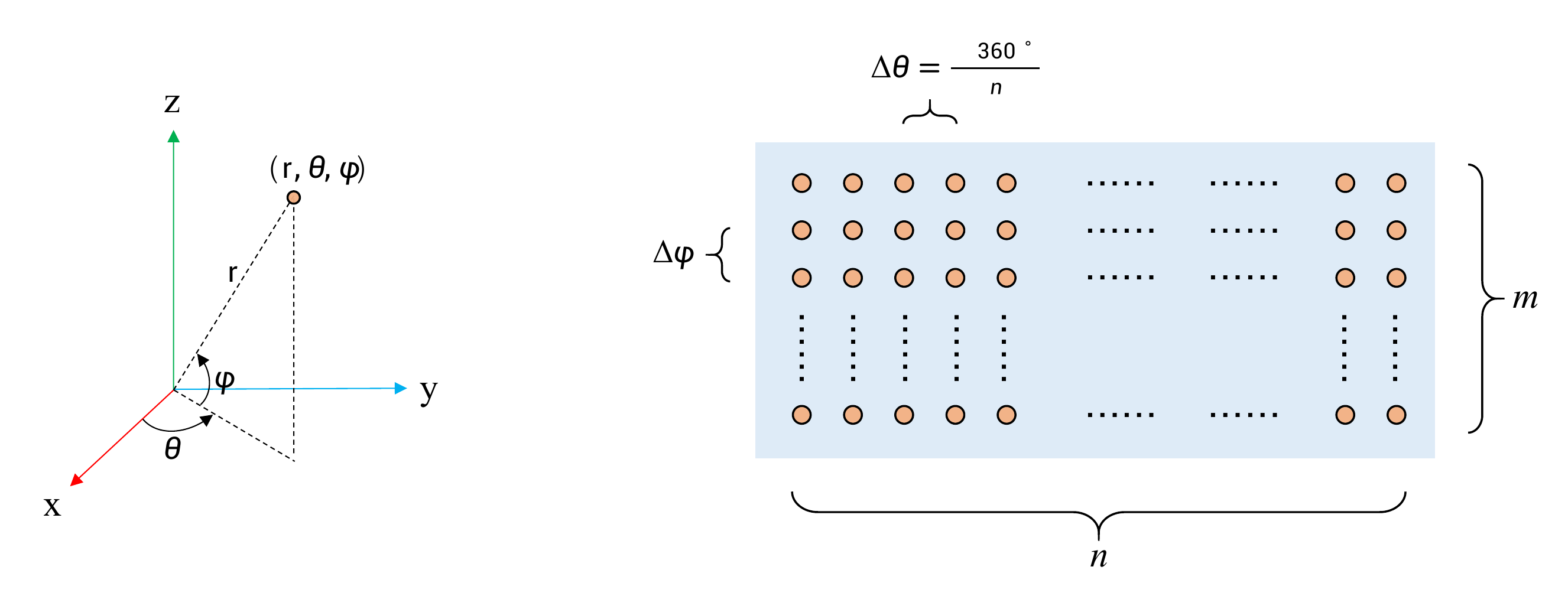}
  \caption{
  \textbf{The range image of the point cloud} (right).  The size of the range image is $m \times n$, where $m$ is the number of beams and $n$ is the number of measurements (\ie, sampling frequency) per scan cycle. A 3D point's coordinates on the range image are computed by discretizing the azimuthal angle $\theta$ and the inclination angle $\phi$ in the spherical coordinate system (left). The $m$, $n$, and $\Delta\phi$ depend on the LiDAR specifications.
  }
  \label{fig:range_view}
\end{figure}

In this work, we advocate a new solution for 3D object detection that works on the range view (RV). As noted by many previous works~\cite{fan2021rangedet, sun2020scalability, meyer2019lasernet}, if we consider these points in the spherical coordinate system and project them in terms of their inclination and azimuth angles, they can form a compact 2D image with size being $m \times n$ (shown in Fig.~\ref{fig:range_view}), where $m$ is the number of beams (\ie, channels) of the LiDAR sensor and $n$ is the sampling frequency per scan cycle. The resulting image is referred to as ``range image'' or ``range view'' of the point cloud. Compared to the aforementioned BEV, the range image is nearly dense and compatible with the LiDAR sampling process, eliminating the need for sparse convolutions and alleviating the loss of points. In addition, the range image closely resembles the common RGB image, minimizing the cost of transferring the 2D detection methods to 3D ones.
In the literature,  %
some works %
attempted 
to detect objects on the range view such as RangeDet~\cite{fan2021rangedet} and LaserNet~\cite{meyer2019lasernet}. These works have shown that RV-based methods can also achieve decent detection performance,
showing the promise. 
However, previous RV-based methods only focus on the single-frame point cloud as the aforementioned range view structure does not hold anymore if the ego (and the origin) moves between the multiple frames. Additionally, the range image of a single-frame point cloud is already nearly dense so that there are not many vacancies the points from other frames can populate. These issues make the range-view based detectors difficult to benefit from the multi-frame fusion. In sharp contrast, the multi-frame fusion can dramatically improve the performance in the BEV-based detectors, as shown in~\cite{yin2021center, lang2019pointpillars}. This makes the performance of RV-based detectors largely lag behind that of the BEV-based ones, hampering their development and application. In this work, we show this issue can be largely remedied with a well-designed Multi-round Range View (MRV) projection mechanism. The proposed MRV makes the RV-based detectors be able to enjoy the gain of multi-frame fusion and thus achieve competitive performance with multi-frame BEV-based detectors.

Here, we summarize our main contributions as follows.
\begin{itemize}
\itemsep -0.2cm
    \item We propose a fully convolutional one-stage 3D object detector, termed \Ours. \Ours\ works on the LiDAR range images and minimizes the gap between the 3D and 2D detectors, while being substantially simpler than current mainstream BEV-based 3D detectors~\cite{yin2021center, zhou2018voxelnet}.
    \item Compared to previous BEV-based detectors~\cite{zhou2018voxelnet, yin2021center}, \Ours sidesteps the complicated voxelization process and eliminates the need for sparse convolutions due to the highly compact range view representation of the point cloud. In the setting of only using the single-frame point cloud as inputs, \Ours can outperform the state-of-the-art BEV-based detector CenterPoint~\cite{yin2021center} while being significantly faster.
    \item We also present a well-designed Multi-round Range View (MRV) projection mechanism, making RV-based detectors %
    be able to 
    benefit from the multi-frame fusion of point clouds as well, and achieve competitive performance compared to the multi-frame BEV-based detectors. To our knowledge, we are the first one approaching the challenge of the RV-based multi-frame point clouds fusion and showing that RV-based detectors can also be boosted by the multi-frame fusion of point clouds.
    \item We believe that our excellent performance of the RV-based detector can be a strong evidence that RV-based detectors compare favorably against the mainstream BEV-based detectors and encourage the community to pay attention to this promising solution.
\end{itemize}

\section{Related Work}
\textbf{Bird-view based 3D Detection.} Most top-performing LiDAR-based 3D detectors~\cite{fan2021embracing, yan2018second, yin2021center, lang2019pointpillars, zhou2018voxelnet} fall into this category, which first convert the point cloud into BEV images. VoxelNet~\cite{zhou2018voxelnet} is the first end-to-end BEV-based detector, which employs PointNet~\cite{qi2017pointnet} to handle the representation within a voxel and 3D convolutions to generate high-level features for the region proposal network (RPN). SECOND \cite{yan2018second} proposes to use sparse convolutions, which can save the computing burden of 3D convolutions. Another popular approach is to eliminate the voxelization along the elevation axis and convert the point cloud into the pillars~\cite{lang2019pointpillars}. Based on the voxel-based or pillar-based BEV representation, CenterPoint~\cite{yin2021center} achieves state-of-the-art performance by using the anchor-free pipelines.

\textbf{Range-view based 3D Detection.} Due to the compactness of the RV representation, some methods~\cite{sun2021rsn, bewley2021range, meyer2019lasernet, meyer2019lasernet} also attempt to perform detection based on the representation. VeloFCN \cite{li2016vehicle} is the pioneering work to perform 3D objection using the range view, which transforms point cloud to the range image and then applies 2D convolution to detect 3D objects. After that, some following works~\cite{meyer2019lasernet, fan2021rangedet} are proposed to narrow the performance gap between RV-based and BEV-based detectors. LaserNet \cite{meyer2019lasernet} models the distribution of 3D box corners to capture their uncertainty, resulting in more accurate defections. RCD~\cite{bewley2021range} introduces the range-conditioned dilation mechanism to dynamically adjust the dilation rate in terms of the measured range, which can alleviate the scale-sensitivity issue of the RV-based detectors. RangeDet~\cite{fan2021rangedet} further proposes the Range Conditioned Pyramid to mitigate the scale-variation issue and utilizes the Meta-Kernel convolution to better exploit the 3D geometric information of the points. To our knowledge, these existing RV-based detectors only take into consideration the single frame point cloud and neglect the substantial improvements brought by the multi-frame fusion as shown in BEV-based detectors.

\section{Our Approach}

\subsection{Range View Representation}\label{sec:range_view}

Given a LiDAR point $(x, y, z)$ in the Cartesian coordinate system with the $z$-axis pointing upward, it can be uniquely transformed to the spherical coordinates $(r, \theta, \phi)$ with
\begin{equation}
    r = \sqrt{x^2 + y^2 + z^2},\ \theta = \atan2(y, x),\ \phi = \atan2(z, \sqrt{x^2 + y^2}),
\end{equation}
where $r$, $\theta$, and $\phi$ are the range, azimuthal angle, and inclination angle, respectively, as shown in Fig.~\ref{fig:range_view}(left). The LiDAR samples the points with a fixed number of beams (denoted by $m$), each of which has a fixed inclination angle. These LiDAR beams synchronously rotate around the $z$-axis uniformly to obtain a $360^\circ$ horizontal field of view and the LiDAR measures a certain number of times (denoted by $n$) per scan cycle (\ie, per frame). Thus, the difference of adjacent measurements' azimuthal angles is $\nicefrac{360}{n}^{\circ}$. For example, on the nuScenes dataset, the LiDAR measures $n = 1086$ times per scan cycle and has $m = 32$ beams. The inclination angles of these beams are evenly spaced from $-30.67^\circ$ to $10.67^\circ$, inclusive. Note that the inclination angles are not always evenly spaced and subject to the specifications of the LiDAR.

Given the regularity of the azimuthal and inclination angles for a given LiDAR, we can discretize the azimuthal angles with $n$ bins, and inclination angles with $m$ bins, respectively. Let $(i, j)$ be the indices of the bins for azimuthal and inclination angles, respectively. By computing all the pairs of $(i, j)$ of the points in a single scan cycle, we can fill these points into a 2D image $I \in \R^{m \times n \times C}$ (\ie, the range image), where $C=9$ consists of the original Cartesian coordinates $(x, y, z)$, the spherical coordinates $(r, \theta, \phi)$, the reflected intensity $i$, the existence $e$ of the point, and a relative timestamp $t$. The existence denotes whether or not the location is filled by a point, and the relative timestamp is only valid in the multi-frame point cloud inputs and denotes the time difference between the frame containing this point and the current frame. In practice, the vehicle itself is often in motion, and this causes that some points might be projected to the same bins on the range image. In this case, we keep the one with the minimal distance to the vehicle.

\subsection{Multi-round Range View Projection (MRV)}\label{sec:mrv}
The point cloud of a single frame is often sparse in the 3D space and of low resolutions. In order to improve the detection performance, the current frame is often combined with several previous frames as the network's inputs~\cite{yin2021center, zhou2018voxelnet}. Taking the nuScenes dataset as an example, the methods on this dataset often take as the inputs $10$ frames, which is composed of the current frame and previous $9$ frames, including $\sim$240K points in total. The crucial issue of the range view representation is the collision that multiple points fall into the same bin happens much more frequently in the multi-frame case. For instance, on nuScenes, only $\sim$28K points are finally kept in the range image and $\sim$90\% of the points are discarded due to the collision. The decimation makes the multi-frame point cloud have almost the same number of valid points with the single-frame version, which is the dominant reason that RV-based methods cannot enjoy the benefit of multi-frame inputs.

In order to cope with the crucial issue, we propose the Multi-round Range View (MRV) projection mechanism. To be specific, we first project the points with the method in Sec.~\ref{sec:range_view}. Then, instead of discarding the rejected points, we project them again with the same process and put the them in another group of nine channels. This projection process is repeated until a sufficient number of points are kept. On the nuScenes dataset, this is repeated five times and the percentage of the kept points can be improved from $\sim$10\% to more than 50\%. In theory, as long as we continue the process, all of the points can be kept. However, we found that the performance is saturated after 5 repetitions on the nuScenes dataset. Finally, the resulting range images of the five rounds are concatenated along the channel dimension and used as the inputs.
Additionally, one caveat is that the points of the current frame should have the highest priority wherever the collision happens because the points of previous frames are stale and might not reflect the current status of the world. Despite being a very simple treatment, it significantly affects the effectiveness of the multi-frame fusion as shown in our experiments.

\subsection{Modality-wise Convolutions}
As mentioned before, each pixel on the single-frame range image contains nine channels. Different the RGB image, whose three channels are of the same modality (\ie, in the color space) and correlated for the manifold of natural images, the nine channels of the range image are not like that and belong to five modalities, \ie, $[x, y, z]$, $[r, \theta, \phi]$, $[i]$, $[e]$ and $[t]$, respectively, where the channels in the one pair of square brackets are of the same modality. For example, the correlation between the reflected intensity $i$ and the coordinate $x$ does not make sense. Further, even the different channel types in the same modality are orthogonal and less correlated as well. For instance, it is difficult to say there is a relationship between the azimuthal angle $\theta$ and the range $r$ of a point. As a result, one channel type should be viewed as an individual ``modality''. By default, the conv.\ layer simultaneously compute spatial correlations and cross-channel correlations. However, as shown before, the channels of the range image are less correlated, and thus, it is not reasonable to use the default conv.\ layer here, and the channels of the range image should be processed separately. This can be easily implemented with the grouped convolutions. Here, we term it \emph{modality-wise convolution} because it is based on the ``modalities''. Once the high-level semantic features of these modalities are individually obtained, we can aggregate the features of these modalities with a $1\times1$ conv.\ layer (\ie, point-wise convolution) for further abstract analysis.

\begin{figure}[t!]
  \centering
  \includegraphics[width=\textwidth]{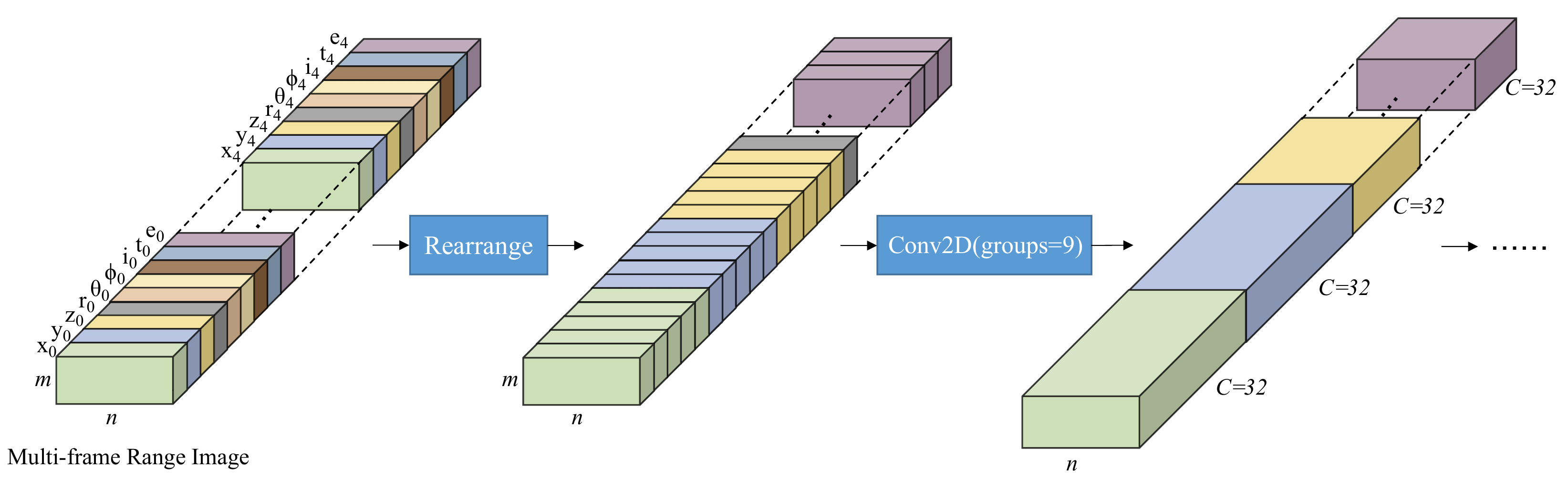}
  \caption{
  \textbf{Modality-wise convolutions with a multi-frame range image.} As we can see, we first rearrange the channels of the multi-frame range image so that the channels with the same type from different frames (\eg, $x_0$, $x_1$, ... $x_{T-1}$) are adjacent. Then, two successive 2D conv.\ layers with the number of groups being 9 (which is equal to the number of different channel types) are used to process these channels separately, mapping each channel type to a 32-channel features. Finally, the features of these channel types are merged by a $1 \times 1$ conv.\ layer.
  }
  \label{fig:modality_wise}
\end{figure}

In this way, the modality-wise convolution is analogous to the widely-used depth-wise convolution, but we highlight that the underlying nature is distinct. Our modality-wise convolution instantiates a reasonable \emph{inductive bias} based on the prior that different modalities are less relevant, and thus it is expected to simultaneously improve both the effectiveness and efficiency, as shown in our experiments. This also leads to the fact that the modality-wise convolution can only be placed at the beginning of the network, where the channels are interpretable and have an explicit modality. In contrast, the depth-wise convolution (together with the point-wise convolution) is often viewed as an efficient approximation of the full conv.\ layer and thus it does not usually yield improved performance and can appear anywhere. Lastly, when it comes to the multi-frame point clouds, the channels of the same type from different frames should be handled together because they are closely correlated. The whole procedure of the modality-wise convolution is illustrated in Fig.~\ref{fig:modality_wise}.

\subsection{Overall Architecture}\label{sec:over_arch}
\begin{figure}[t!]
  \centering
  \includegraphics[width=\textwidth]{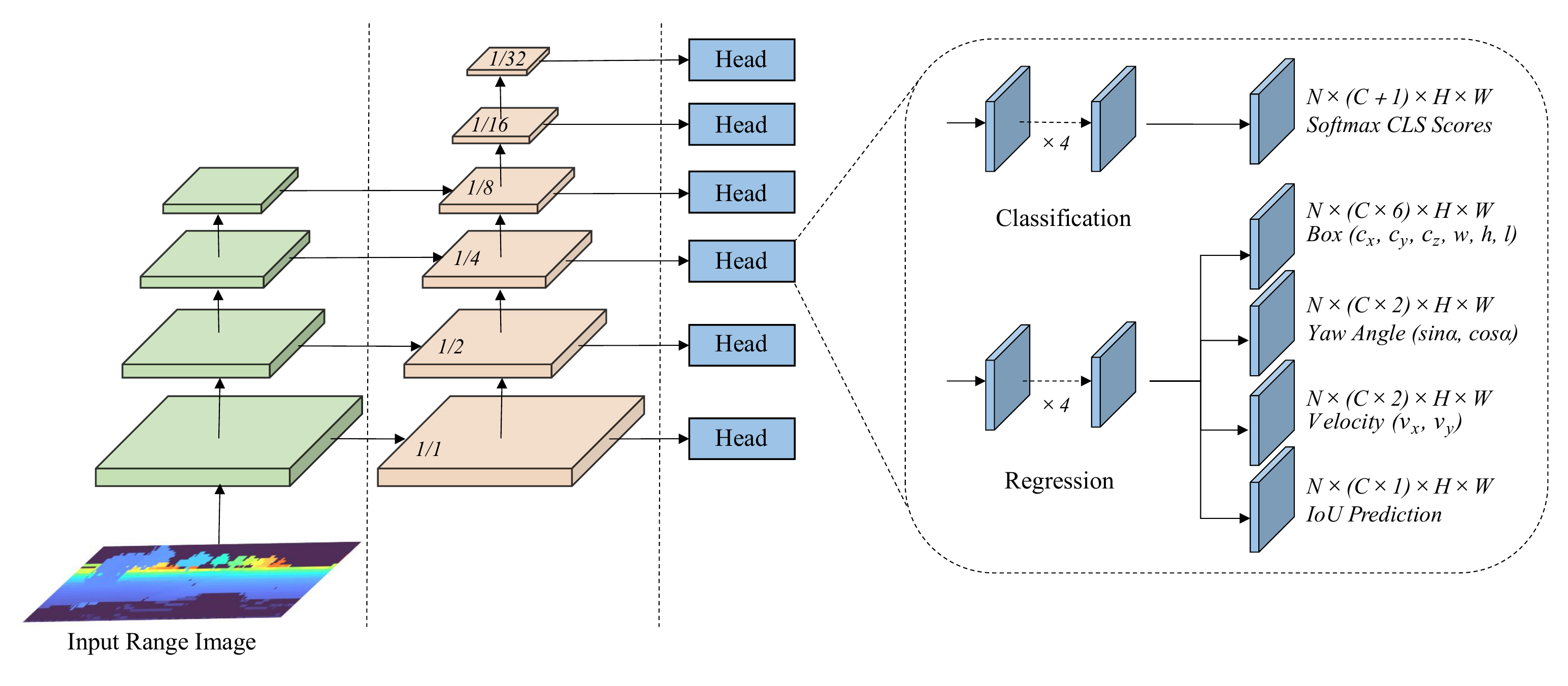}
  \caption{
  \textbf{Overall architecture.} The overall architecture of \Ours\ resembles the 2D image-based detector FCOS~\cite{tian2019fcos}. By taking as input 
  an 
  range image, the network obtains the multi-level FPN features, and then the classification and regression branches are attached to these feature levels to predict the final 3D boxes. Different from FCOS, the weights of the detection heads are not shared between the FPN levels as mentioned in Sec.~\ref{sec:over_arch}. In addition, the class-specific regression heads are used instead of the class-agnostic ones in FCOS.
  }
  \label{fig:overall_arch}
\end{figure}

The overall architecture of \Ours is shown in Fig.~\ref{fig:overall_arch}. \Ours\ follows the spirit of the anchor-free detector FCOS~\cite{tian2019fcos} in the image-based object detection and is a standard fully convolutional network~\cite{long2015fully}.

Taking a (multi-frame) range image $I \in \R^{m \times n \times (T \times C)}$ as an example, where $T$ is the number of the used frames, we first forward the range image through our backbone network, termed LiDAR-Net. LiDAR-Net is adapted from ResNet-50~\cite{he2016deep}. Specifically, before the first conv.\ layer of ResNet-50, we insert three branches of the modality-wise convolutions with three dilation rates being $1$, $3$, and $6$, respectively. The branches with various dilation rates aim to capture the multi-scale context, whose outputs are summed up. Then, the first two $2\times$ downsamplings of ResNet-50 are removed, which is of great importance due to the low resolutions of the range images. Moreover, we change the numbers of blocks of ResNet-50's four stages from $(3, 4, 6, 3)$ to $(4, 4, 1, 1)$ and stop doubling the number of channels in the third and fourth stages because we found that the final performance is not sensitive to the capacity of the later stages but quite sensitive to that of the early stages. This ravels one of the important difference between the LiDAR-based and image-based object detection tasks. For object detection in RGB images, more convolutions in the later stages are often required to transform the RGB pixels into the highly semantic and abstract features that can be used to obtain the geometries of the objects. However, the LiDAR points themselves are already geometric points and thus that many convolutions in the later stages are no longer needed. Thus, we can instead allocate the capacity in the later stages to the early stages (before any downsampling) to better incorporate the geometric information carried by the raw points. Finally, the four levels of feature maps from LiDAR-Net's four stages are used, denoted by $C_2$, $C_3$, $C_4$, and $C_5$, respectively.

Next, following FCOS, the four levels of feature maps are sent into a feature pyramid network (FPN)~\cite{lin2017feature} to obtain six levels of pyramid feature maps denoted by $P_2$, $P_3$, $P_4$, $P_5$, $P_6$, and $P_7$. Their spatial downsampling ratios to the input are $\nicefrac{1}{1}$, $\nicefrac{1}{2}$, $\nicefrac{1}{4}$, $\nicefrac{1}{8}$, $\nicefrac{1}{16}$, and $\nicefrac{1}{32}$, respectively. Then, similar to FCOS, the classification and regression heads, each with four $3 \times 3$ conv.\ layers of channel 64 and a final prediction conv.\ layer (for the classification or regression), are attached to these feature levels. Unlike the heads in image-based detectors, whose weights are shared between these feature levels, it is important in the LiDAR-based detector to untie these weights. This is another important difference between image-based and LiDAR-based detectors. In image-based detectors, different sizes of objects can be normalized to similar sizes of objects by the downsampling the image with corresponding factors. That is what the ``pyramid'' means in the FPN. Thus, the image-based detectors can share the weights of the detection heads because the objects have been normalized to similar sizes after the FPN. However, in LiDAR-based detection, the objects' sizes cannot be normalized in this way because the sizes of the objects are determined by the 3D points' coordinate values in them and downsampling the range images cannot alter their real sizes in the 3D space. Therefore, it is no longer reasonable to share the detection heads between these FPN levels. This is also confirmed in our experiments.

\textbf{Training Targets Computation.} Similar to FCOS~\cite{tian2019fcos}, we need to assign the training targets to each pixel on the range image. The training targets computation is described using the nuScenes dataset as an example. On nuScenes, an object's ground-truth 3D box is parameterized by $(c_x^*, c_y^*, c_z^*, w^*, h^*, l^*, \alpha^*)$, where $(c_x^*, c_y^*, c_z^*)$ is the 3D center of the box, and $w$, $h$, $l$, and $\alpha$ respectively are the width, height, length, and the yaw angle around the $z$ axis. If the original 3D point of a pixel on the range image is contained in the 3D box of an object, the pixel is responsible for the object and predicts its category, 3D box and etc.. Here, the 3D box regression targets of the pixel are relative to the original 3D coordinates of the range image pixel and are defined as
\begin{equation}
    \Delta c_x = c_x^* - x_0,\ \Delta c_y = c_y^* - y_0,\ \Delta c_z = c_z^* - z_0,\ t_w = \log(w^*),\ t_h = \log(h^*),\ t_l = \log(l^*),
\end{equation}
where $(x_0, y_0, z_0)$ are the original 3D coordinates of the range image pixel. Similarly, the regression targets of the yaw angle $\alpha$ are also relative to the azimuthal angle of the pixel, and following the convention~\cite{yin2021center}, we decouple the relative azimuthal angle into $(\sin\Delta\alpha, \cos\Delta\alpha)$. On nuScenes, we also need to predict the velocity vector $(v_x^*, v_y^*)$ of the object, which are used as the training targets as is. Other pixels whose 3D points are not in any 3D box are used as the negative samples. Note that the points within a object's 3D box form a 2D mask on the range image.

\textbf{Network Outputs.} The nuScenes dataset has $C = 10$ classes of interest, which is predicted by the classification head followed by a $\softmax$ layer (the upper head in Fig.~\ref{fig:overall_arch}). The other regression targets are predicted by four sibling output heads of the regression branch, respectively, as shown in Fig.~\ref{fig:overall_arch}. Here, we use the class-specific regression predictions and thus the number of the output channels is amplified by $C$ times. Moreover, following~\cite{ge2021yolox, redmon2016you}, each positive pixel also predicts the intersection-over-union (IoU) between the predicted 3D box and ground-truth one, which is multiplied to the classification score before the non-maximum suppression (NMS).

\textbf{Loss Functions.} The classification predictions are supervised with the cross entropy (CE) loss. Following ~\cite{ge2021ota}, we dynamically reassign the classification labels during training. Specifically, for a ground-truth 3D box, only the top $K$ pixels whose predicted 3D boxes have the lowest costs with the ground-truth box are assigned with the positive labels and other pixels are considered negative, where the cost is defined as the summation of the classification loss and the opposite of the IoU between the predicted boxes and the ground-truth box, and $K$ is dynamically calculated, being the summation (rounded to an integer) of the highest $Q = 20$ IoUs between the predicted boxes and the ground-truth box. For the 3D box regression, we make use of both the IoU loss~\cite{yu2016unitbox} and $L_1$ loss. Following ~\cite{tian2019fcos}, the IoU predictions are penalized with the binary cross entropy (BCE) loss since they are in the range $[0, 1]$.

\textbf{Inference.} The inference is very similar to that of the 2D image-based detector FCOS. To be specific, the range images are forwarded through the network and the aforementioned predictions are obtained. The predictions are filtered in terms of the classification scores and only the predictions with the score greater than $0.01$ are kept. The 3D boxes are restored by inverting the computation of the training targets. Then, the NMS on the 3D boxes is applied with threshold $0.2$ to remove the duplications. Finally, the top $500$ predictions with the highest scores are used as the final predictions.

\section{Experiments}
We conduct experiments on the nuScenes dataset~\cite{caesar2020nuscenes}, which contains 1000 scenes with 700, 150, and 150 scenes for training, validation, and testing, respectively. The training, validation, and testing sets have 28K, 6K, 6K keyframes annotated with 10 classes, respectively. For all ablation experiments, we train the models on the training set and report the performance on the validation set unless specified. The metrics of the 3D detection task are mean Average Precision (mAP) and the nuScenes detection score (NDS). The LiDAR sensor of the nuScenes dataset has $m = 32$ beams and $n = 1086$ measurements per scan cycle. The mAP is based the center distances on the bird-eye view at thresholds 0.5m, 1m, 2m, 4m in place of the box IoUs. NDS is a weighted average of mAP, the translation error, the scale error, the orientation error, the velocity error, and the box attributes error.

\textbf{Implementation Details.} Unless specified, following~\cite{yin2021center}, \Ours\ is trained by 40 epochs with the AdamW~\cite{loshchilov2017decoupled} optimizer under the MMDetection3D framework~\cite{mmdet3d2020}, which takes $\sim$26 hours on 8 A100 GPUs.
 The one-cycle learning rate policy~\cite{smith2019super} with initial learning rate %
 $ 10^{-3}$ is used.
 The learning rate gradually increases to %
 $ 0.01 $
 in the first 40\% epochs and then gradually decreases to %
 $ 10^{-7}$
 in the rest of the training process. The weight decay is 0.01, and the momentum ranges from 0.85 to 0.95. In addition, due to the low vertical resolution of the range image on the nuScenes dataset, we upscale the range image in the vertical direction by $2$ with the nearest interpolation. During training, the point cloud is randomly flipped along both the 
 $x$ and $y$ axes and rotated in the range [$-\pi$, $\pi$], as well as globally scaled by a random factor from 
 $[0.95, 1.05]$. The ground-truth copy-paste data augmentation from ~\cite{yan2018second} is also used. For multi-frame point cloud, we use 10 sweeps in total, as in previous works~\cite{yin2021center, lang2019pointpillars}. The inference time is measured on a 3090Ti GPU with batch size 1.

\subsection{Multi-round Range View Projection}
Here, we conduct experiments to demonstrate the effectiveness of the proposed multi-round range view (MRV) projection. As shown in Table~\ref{table:multi_round}, in the single-frame settings, MRV is also helpful, for example, by increasing the number of rounds from $1$ to $3$, the mAP can be boosted by $0.62\%$. This is due to the fact that the vehicle is often in motion and the aforementioned collisions happen within a single frame as well. As you can see, if one round is used, the multi-frame fusion can improve the mAP by a substantial margin $\sim$1.9\% (from 53.14\% to 55.02\%). The improvement can be dramatically increased to 3.9\% mAP if we use 5 rounds in MRV (57.8\% mAP), due to the fact that 5-round MRV can keep much more points in the multi-frame point clouds as mentioned in Sec.~\ref{sec:mrv}. This confirmed the effectiveness of MRV. Note that elapsed time of MRV is insensitive to the number of rounds as shown in Table~\ref{table:multi_round}.
\begin{table}[t!]
\caption{
\textbf{Multi-round range view (MRV) projection.} Time: the elapsed time of MRV.}
  \label{table:multi_round}
  \setlength{\tabcolsep}{14pt}
  \centering
  \begin{tabular}{cccccc}
    \Xhline{2\arrayrulewidth}
    & & \multicolumn{2}{c}{Single-frame} & \multicolumn{2}{c}{Multi-frame} \\
    \#Rounds & Time (ms) &  mAP (\%) & NDS (\%) & mAP (\%) & NDS (\%) \\
    \hline
    1 & \textbf{0.66} & 53.14 & 51.52 & 55.02 & 61.27 \\
    3 & 0.72 & \textbf{53.76} & \textbf{53.62} & 56.01 & 62.82 \\
    5 & 0.76 & 53.42 & 53.40 & \textbf{57.08} & 63.15 \\
    7 & 0.76 & 53.06 & 53.10 & 56.99 & \textbf{63.47} \\
    \Xhline{2\arrayrulewidth}
  \end{tabular}
\end{table}
\begin{table}
\caption{
\textbf{Whether to make the points of the current frame have the highest priority.} ``C.V.'',  ``Ped.'' and ``C.T.'' indicate ``construction vehicle'', ``pedestrian'' and ``traffic cone'', respectively.}
  \label{table:priority}
  \setlength{\tabcolsep}{2.2pt}
  \centering
  \begin{tabular}{cccccccccccccc}
    \Xhline{2\arrayrulewidth}
    Prioritized? & mAP(\%) & NDS(\%) & Car & Truck & Bus & Trailer & C.V. & Ped. &  Motor & Bicycle & T.C. & Barrier \\
    \hline
     & 54.29 & 62.31 & 76.8 & 47.3 & 62.0 & 32.0 & 16.5 & 80.3 & 53.8 & \textbf{38.2} & 70.1 & 65.9 \\
    \checkmark & \textbf{57.08} & \textbf{63.15} & \textbf{82.1} & \textbf{52.3} & \textbf{65.2} & \textbf{33.6} & \textbf{18.3} & \textbf{84.1} & \textbf{58.5} & 35.3 & \textbf{73.4} & \textbf{67.9} \\
    \Xhline{2\arrayrulewidth}
  \end{tabular}
\end{table}

More importantly, in the RV-based multi-frame settings, it is crucial to assign the highest priority to the points from the current frame when the collision happens. As shown in Table~\ref{table:priority}, without this treatment, the performance of the multi-frame fusion is significantly dropped from $57.08\%$ to $54.29\%$ in mAP. Note that the single-frame counterpart can already achieve mAP 53.42\%. We argue that the neglect of this point in previous works is one of the main reasons that RV-based detectors can barely enjoy the benefit of multi-frame fusion.

\subsection{Modality-wise Convolutions}
\begin{table}[h!]
\caption{
\textbf{Modality-wise convolutions.} ``Multi-frame'': whether to group together the channels with the same type from multiple frames. Time: the latency of this module.}
  \label{table:modality_wise}
  \setlength{\tabcolsep}{10pt}
  \centering
  \begin{tabular}{ccccc}
    \Xhline{2\arrayrulewidth}
    Modality groups & Multi-frame & Time (ms) & mAP (\%) & NDS (\%) \\
    \hline
    $[x, y, z, r, \theta, \phi, i, t, e]$ & \checkmark & 6.0 & 56.35 & 62.65 \\
    $[x, y, z], [r, \theta, \phi], [i], [t], [e]$ & \checkmark & 4.2 & 56.34 & 63.07 \\
    $[x], [y], [z], [r], [\theta], [\phi], [i], [t], [e]$ & \checkmark & \textbf{3.2} & \textbf{57.08} & \textbf{63.15} \\
    $[x], [y], [z], [r], [\theta], [\phi], [i], [t], [e]$ & & 3.8 & 56.83 & \textbf{63.15} \\
    \Xhline{2\arrayrulewidth}
  \end{tabular}
\end{table}
\begin{table}[h]
\caption{
\textbf{Multi-scale context aggregation.}}
  \label{table:multi_scale}
  \setlength{\tabcolsep}{2.55pt}
  \centering
  \begin{tabular}{cccccccccccccc}
    \Xhline{2\arrayrulewidth}
    Dilations & mAP(\%) & NDS%
    (\%) & Car & Truck & Bus & Trailer & C.V. & Ped. &  Motor & Bicycle & T.C. & Barrier \\
    \hline
    $(1,)$ & 55.87 & 62.68 & 81.8 & 51.0 & 64.1 & 32.6 & 16.9 & 83.8 & 56.3 & 31.2 & 72.7 & \textbf{68.3} \\
    $(1, 3)$ & 56.39 & 62.96 & \textbf{82.3} & 52.1 & \textbf{65.3} & 32.8 & 17.1 & 83.7 & 58.2 & 32.5 & 72.0 & 67.8 \\
    $(1, 3, 6)$ & \textbf{57.08} & \textbf{63.15} & 82.1 & \textbf{52.3} & 65.2 & \textbf{33.6} & \textbf{18.3} & \textbf{84.1} & 58.5 & \textbf{35.3} & \textbf{73.4} & 67.9 \\
    $(1, 1, 1)$ & 56.63 & 63.12 & 82.1 & 51.5 & 65.2 & 32.9 & 17.1 & 83.7 & \textbf{58.8} & 33.7 & 73.1 & 68.0 \\
    \Xhline{2\arrayrulewidth}
  \end{tabular}
\end{table}
The experimental results of our modality-wise convolutions are shown in Table~\ref{table:modality_wise}. If we use the full convolutions here, which compute the correlation between all the channels of the range image, \Ours\ can achieve 56.35\% in mAP. By splitting the channels into various modalities (2nd row), the performance can be slightly improved, \ie, from NDS 62.65\% to 63.07\%. Moreover, due to the fact that even the different channel types of the same modality are orthogonal and less correlated, we process each channel type individually. As shown in the table, the performance can be further boosted from mAP 56.34\% to 57.08\% (3rd row) while the lowest latency is achieved. Finally, the last row shows the results if we do not consider the channels with the same type from multiple frames together, where the mAP is slightly worse.

In addition, as mentioned before, we employ multiple modality-wise convolution branches with various dilation rates in parallel to capture the multi-scale context. The experiments are shown in Table~\ref{table:multi_scale}. 
As %
we 
can see, compared with the one using only one dilation rate, using multiple dilation rates can improve the performance by more than $1\%$ in mAP (from 55.87\% to 57.08\%).
\subsection{%
Untied 
Weights of Detection Heads} 
\begin{table}[h!]
\caption{
\textbf{Whether to untie the weights of the detection heads.}}
  \label{table:shared_heads}
  \setlength{\tabcolsep}{2.75pt}
  \centering
  \begin{tabular}{cccccccccccccc}
    \Xhline{2\arrayrulewidth}
    Untied? & mAP(\%) & NDS(\%) & Car & Truck & Bus & Trailer & C.V. & Ped. &  Motor & Bicycle & T.C. & Barrier \\
    \hline
    & 56.44 & 63.09 & 82.0 & 51.1 & 64.3 & 31.1 & \textbf{18.3} & 83.8 & 57.9 & 34.7 & 72.9 & \textbf{68.4} \\
    \checkmark & \textbf{57.08} & \textbf{63.15} & \textbf{82.1} & \textbf{52.3} & \textbf{65.2} & \textbf{33.6} & \textbf{18.3} & \textbf{84.1} & \textbf{58.5} & \textbf{35.3} & \textbf{73.4} & 67.9 \\
    \Xhline{2\arrayrulewidth}
  \end{tabular}
\end{table}
As mentioned before, it is better to untie the weights of the detection heads between the FPN levels in the LiDAR-based detector. This is confirmed in Table~\ref{table:shared_heads}. As we can see, by untying the weights, the performance can be improved from mAP 56.44\% to 57.08\%. This ravels one of the interesting differences between image-based and LiDAR-based detectors because the shared detection heads between FPN levels often achieve better performance in image-based detectors~\cite{tian2019fcos, lin2017feature}.
\begin{table}[h!]
\caption{
\textbf{Inference time breakdowns.} We compare against 
the state-of-the-art BEV-based CenterPoint \cite{yin2021center}, which is trained with exactly the same strategies. \Ours\ is significantly faster as well as competitive in the multi-frame setting (and superior in the single-frame setting).}
  \label{table:inference_time}
  \setlength{\tabcolsep}{0.75pt}
  \centering
  \begin{tabular}{lcccccc}
    \Xhline{2\arrayrulewidth}
    Method & Voxel/MRV(ms) & Backbone%
    (ms) & Heads(ms) & Overall(ms) & mAP(\%) & NDS(\%) \\
    \hline
    \textit{single-frame point cloud} &&&&& \\
    CenterPoint~\cite{yin2021center} & 1.62 & 41 & 8 & 50.62 & 52.83 & \textbf{53.86} \\
     {\Ours} (\textbf{Ours}) & \textbf{0.63} & \textbf{31} & \textbf{7} & \textbf{38.63} & \textbf{53.42} & 53.40 \\
    \hline
    \textit{multi-frame point cloud} &&&&& \\
    CenterPoint~\cite{yin2021center} & 1.95 & 63 & 9 & 73.95  & \textbf{60.40} & \textbf{67.25} \\
     {\Ours} (\textbf{Ours}) & \textbf{0.76} & \textbf{31} & \textbf{7} & \textbf{38.76} & 57.08 & 63.15 \\
    \Xhline{2\arrayrulewidth}
  \end{tabular}
\end{table}

\subsection{Inference Time Comparisons}
We compare the inference time of \Ours\ and the state-of-the-art BEV-based detector CenterPoint~\cite{yin2021center}. As shown in Table~\ref{table:inference_time}, the proposed MRV is faster than the voxelization in CenterPoint. It is worth noting that the voxelization algorithm reported here is
 \emph{nondeterministic}, which cannot yield stable results but being significantly faster. In the official code of MMDetection3D~\cite{mmdet3d2020}, the deterministic voxelization takes $\sim$100ms for the multi-frame point clouds. In sharp contrast, the proposed MRV is always \emph{deterministic} and highly efficient. Moreover, due to the compactness of the range image, our network can be implemented with the standard convolutions alone, thus being much more efficient than CenterPoint using the sparse convolutions as shown in the table. The elapsed time of the post-processing is omitted here as it is closely similar in both methods.

\subsection{Comparisons with State-of-the-art Methods}
We further compare \Ours\ with other state-of-the-art methods on the nuScenes test set. For the model on the test set, we increase the number of channels in the detection heads from 64 to 128, which improve the validation mAP by $\sim$0.6\% with slightly longer latency. Additionally, we remove the copy-paste data augmentation in the last 5 epochs during training as in~\cite{wang2021pointaugmenting}. As shown in Table~\ref{table:sota}, \Ours\ achieves competitive performance with other state-of-the-art methods.
\begin{table}[t!]
\caption{
\textbf{Comparisons with state-of-the-art methods on the nuScenes test set.} The results are directly %
quoted 
from their original papers.}
  \label{table:sota}
  \setlength{\tabcolsep}{0.65pt}
  \centering
  \begin{tabular}{rccccccccccccc}
    \Xhline{2\arrayrulewidth}
    Method & mAP(\%) & NDS(\%) & Car & Truck & Bus & Trailer & C.V. & Ped. &  Motor & Bicycle & T.C. & Barrier \\
    \hline
    PointPillars~\cite{lang2019pointpillars} & 30.5 & 45.3 & 68.4 & 23.0 & 28.2 & 23.4 & 4.1 & 59.7 & 27.4 & 1.1 & 30.8 & 38.9 \\
    SSN \cite{zhu2020ssn} &  46.3 & 56.9& 80.7 & 37.5 & 39.9 & 43.9 &  14.6 &  72.3 & 43.7 & 20.1 &  54.2 &  56.3  \\
    CVCNet~\cite{chen2020every} & 55.3 & 64.4 & 82.7 & 46.1 & 46.6 & 49.4 & 22.6 & 79.8 & 59.1 & 31.4 & 65.6 & 69.6 \\
    CBGS~\cite{yin2021center} & 52.8 & 63.3 & 81.1 & 48.5 & 54.9 & 42.9 & 10.5 & 80.1 & 51.5 & 22.3 & 70.9 & 65.7 \\
    CenterPoint~\cite{yin2021center} & 58.0 & 65.5 & \textbf{84.6} & \textbf{51.0} & \textbf{60.2}& \textbf{53.2} & 17.5 & 83.4 & 53.7 & 23.7 & 76.7 & \textbf{70.9} \\
    \hline
    \Ours (c128) & \textbf{60.2} & \textbf{65.7} & 82.2 & 47.7 & 52.9 & 48.8 & \textbf{28.8} & \textbf{84.5} & \textbf{68.0} & \textbf{39.0} & \textbf{79.2} & 70.7 \\
    \Xhline{2\arrayrulewidth}
  \end{tabular}
\end{table}

\section
{Conclusion}
We have presented an efficient range-view-based 3D object detector \Ours. \Ours shows the challenging LiDAR-based object detection can also be solved with the standard convolutions alone, similar to what we have done in the image-based 2D object detection. We also for the first time show the RV-based 3D detector can also enjoy the benefit of the multi-frame fusion with the proposed MRV. We hope our strong results can encourage the community to pay more attention to this promising direction.

\textbf{Societal Impacts.} This paper presents a method that can locate the 3D location of the objects of interest in LiDAR point clouds. This technique might be abused for military purposes, for example, on lethal autonomous weapons.

\appendix 
\section*{Appendix}  
   
\section{More Experiments}
As mentioned before, for the model on the test set, we use 128 channels in the detection heads, which improve the performance from 57.08\% to 57.71\% mAP with 6ms more latency, as shown in Table~\ref{table:varying_channels}. In Table~\ref{table:varying_modality_wise_conv}, we vary the number of the conv.\ layers in the modality-wise convolutions. As we can see, using two conv.\ layers achieves the best performance here.

In addition, we compare with more popular 3D detectors by keeping the training schedule the same. As shown in Table~\ref{table:sota2}, \Ours\ significantly outperforms these methods.

\begin{table}[t!]
\caption{
\textbf{Varying the number of the channels in the detection heads.} Time: the latency of the detection heads.}
  \label{table:varying_channels}
  \setlength{\tabcolsep}{0.78pt}
  \centering
  \begin{tabular}{cccccccccccccc}
    \Xhline{2\arrayrulewidth}
    \#Channels & Time (ms) & mAP(\%) & NDS(\%) & Car & Truck & Bus & Trailer & C.V. & Ped. &  Motor & Bicycle & T.C. & Barrier \\
    \hline
    64 & \textbf{7} & \textbf{57.08} & 63.15 & 82.1 & 52.3 & 65.2 & 33.6 & \textbf{18.3} & \textbf{84.1} & \textbf{58.5} & \textbf{35.3} & 73.4 & 67.9 \\
    128 & 13 & 57.71 & \textbf{64.09} & \textbf{82.9} & \textbf{53.4} & \textbf{66.5} & \textbf{34.7} & 18.1 & \textbf{84.1} & \textbf{58.5} & 35.2 & \textbf{74.3} & \textbf{69.4} \\
    \Xhline{2\arrayrulewidth}
  \end{tabular}
\end{table}

\begin{table}[h!]
\caption{
\textbf{Varying the number of the conv.\ layers in the modality-wise convolutions.} Time: the latency of the modality-wise convolutions.}
  \label{table:varying_modality_wise_conv}
  \setlength{\tabcolsep}{1.3pt}
  \centering
  \begin{tabular}{cccccccccccccc}
    \Xhline{2\arrayrulewidth}
    \#Conv.\ & Time (ms) & mAP(\%) & NDS(\%) & Car & Truck & Bus & Trailer & C.V. & Ped. &  Motor & Bicycle & T.C. & Barrier \\
    \hline
    1 & \textbf{2.0} & 55.81 & 61.98 & 81.8 & 51.4 & 65.2 & 30.7 & 16.9 & 83.4 & 56.0 & 32.8 & 71.9 & 68.1 \\
    2 & 3.2 & \textbf{57.08} & 63.15 & 82.1 & \textbf{52.3} & 65.2 & \textbf{33.6} & 18.3 & \textbf{84.1} & \textbf{58.5} & \textbf{35.3} & \textbf{73.4} & 67.9 \\
    3 & 4.3 &  56.71 & \textbf{63.25} & \textbf{82.6} & 51.4 & \textbf{65.3} & 31.9 & \textbf{18.5} & 83.9 & 57.3 & 34.9 & 72.9 & \textbf{68.4} \\
    \Xhline{2\arrayrulewidth}
  \end{tabular}
\end{table}

\begin{table}[h!]
\caption{
\textbf{Comparisons with more popular methods on the nuScenes validation set.} To make fair comparisons, the other methods are also trained for 40 epochs using the implementation in MMDetection3D.}
  \label{table:sota2}
  \setlength{\tabcolsep}{1.75pt}
  \centering
  \begin{tabular}{rccccccccccccc}
    \Xhline{2\arrayrulewidth}
    Method & mAP(\%) & NDS(\%) & \rotatebox{60}{Car} &
    \rotatebox{60}{Truck} & \rotatebox{60}{Bus} & \rotatebox{60}{Trailer} & \rotatebox{60}{C.V.} & \rotatebox{60}{Ped.} &  \rotatebox{60}{Motor} & \rotatebox{60}{Bicycle} & \rotatebox{60}{T.C.} & 
    \rotatebox{60}{Barrier} \\
    \hline
    P.Pillars
    \cite{lang2019pointpillars} & 41.3 & 54.9 & 81.3 & 37.7 & 48.5 & 29.9 & 8.5 & 72.7 & 38.9 & 30.6 & 33.5 & 37.2 \\
    SSN \cite{zhu2020ssn} &  45.0 & 57.3& \textbf{82.3} & 44.9 & 59.8 & 29.6 &  12.8 &  69.8 & 47.9 & 23.2 &  24.8 &  40.3  \\
    \hline
    \Ours & \textbf{57.08} & \textbf{63.15} & 82.1 & \textbf{52.3} & \textbf{65.2} & \textbf{33.6} & \textbf{18.3} & \textbf{84.1} & \textbf{58.5} & \textbf{35.3} & \textbf{73.4} & \textbf{67.9} \\
    \Xhline{2\arrayrulewidth}
  \end{tabular}
\end{table}

\section{Visualization}
\begin{figure}[h!]
  \centering
  \includegraphics[width=\textwidth]{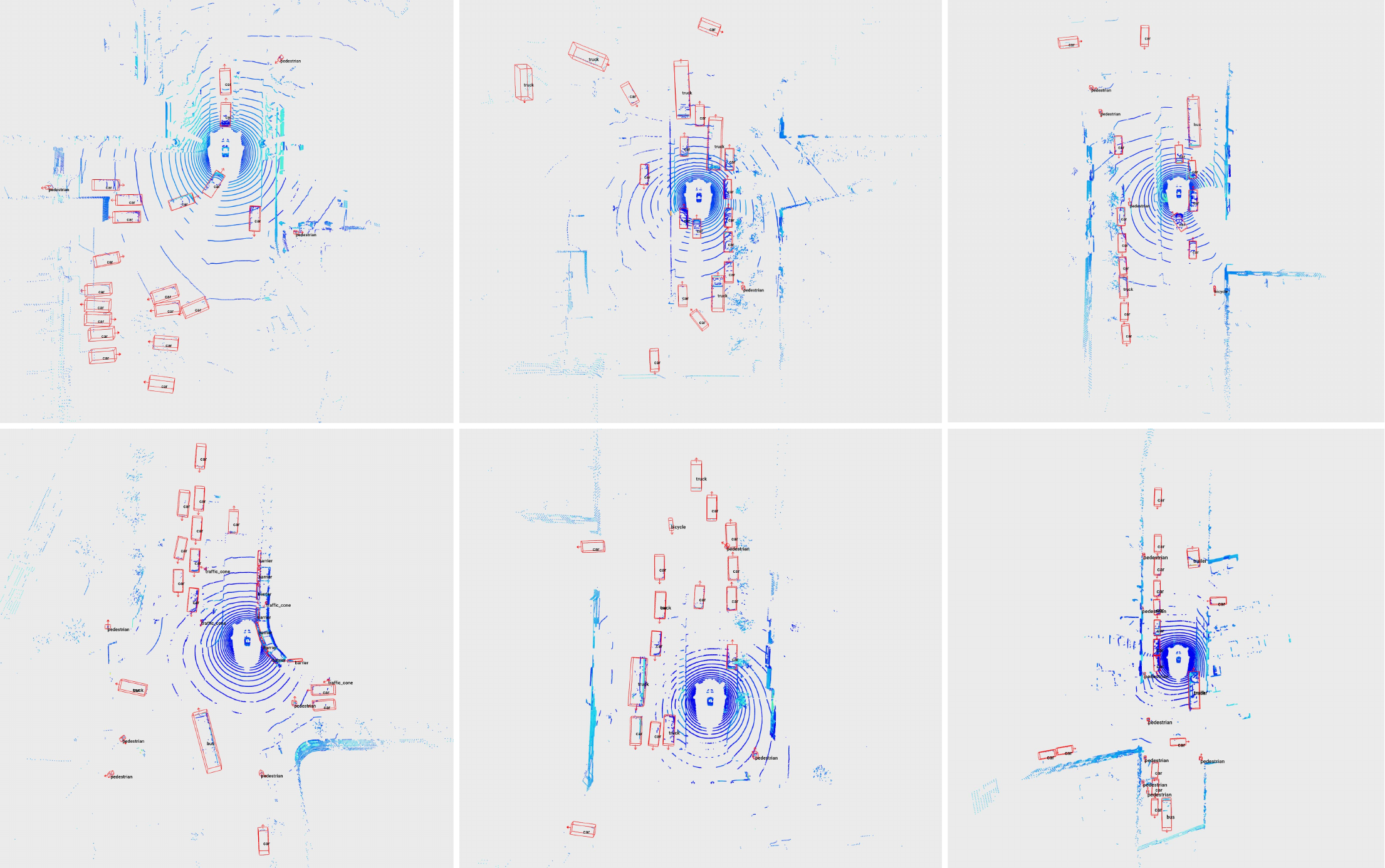}
  \caption{
  \textbf{Visualization results of \Ours on the nuScenes val.\  set.}
  }
  \label{fig:vis}
\end{figure}

Some visualization results are shown in Fig.~\ref{fig:vis}. As we can see, \Ours\ can work reliably under a wide variety of challenging circumstances.

\section{Rejoinder} 

\textbf{Q: Direct comparisons with other range-based 3D detectors.}

None of these RV-based methods releases their full code. This makes it very challenging to compare ours and theirs due to the huge differences in training/testing settings, data augmentation, network architectures, pre-/post-processing etc. In addition, these previous range view detectors are not really fully convolutional, being much more cumbersome than ours. For example, both RCD and RSN are two-stage methods. RCD uses a second stage like RCNN to refine the initial box proposals, and RSN only uses the range view for points filtering in the first stage and still relies on BEV in its second stage. RangeDet exploits the Meta-Kernel technique, which dynamically generates the weights of convolutional kernels, hampering the on-device deployment ({\it e.g.}, TensorRT/ONNX deployment). In contrast, our model is one-stage and only depends on the standard convolutions, thus being significantly simple and easy to deploy on self-driving cars in practice. Moreover, none of the existing range view methods can cope with the multi-frame fusion, which is one of our important contributions and was previously considered intractable in the range view.

We mainly compare our methods with mainstream BEV-based methods such as CenterPoint and PointPillar, showing that an RV-based method compares favourably with these popular BEV-based solutions even in the multi-frame case. These BEV works are used extensively in practice, and thus we think the performance competitive with theirs can demonstrate the effectiveness of our method. Finally, we will release our full code to facilitate the research of RV-based detectors.


\textbf{Q: Different performance compared with CenterPoint on the nuScenes test set and the val.\ set.}

We only use FCOS-LiDAR (C128) on the test set. The model on the val set is smaller and instead has only $64$ channels in its detection head. Moreover, for the experiments on the val set, the training/testing settings are strictly controlled to ensure a fair comparison between ours and CenterPoint. For the model on the test set, we further use the ``fade strateg''  during training ({\it i.e.}, removing the copy-paste data augmentation in the last 5 epochs). This can improve the performance by about $2\%$ mAP. Additionally, the test set results of other methods are directly token from their original papers and there might be other subtle differences in the training/testing process. This is why our method shows better performance than CenterPoint on the test set.

\textbf{Q: The feature map of each level has to be resized to the original image size.}

No, we do {\bf not}  resize the feature maps of all levels to the original image size. 
Only the first level of feature maps has the same size as the original image size, and other levels are down-sampled by powers of $2$, respectively, as in the standard FPN. Thus, FPN is still needed.

\textbf{Q. Does random scale augmentation cause object artifacts?}

Almost not for two reasons. 1) We apply the random scale augmentation globally, i.e., all points in the same point cloud are proportionally scaled by the same scale factor at a time. As a result, this does not alter the azimuth and inclination angles of these points in the spherical coordinates system, and neither do the range view projections of these points. 2) We choose the scale factor in the range from $0.95$ to $1.05$, which only changes the point cloud by a small amount and thus will not cause object artifacts.

\textbf{Q. MRV's performance in multi-frame settings still falls behind BEV's.}

Yes, the multi-frame fusion in range view still is an open question and more efforts are needed in order to catch BEV's. But we would like to highlight that we are the first one to demonstrate that range view detectors can also benefit from multi-frame fusion. The difficulty of multi-frame fusion is previously considered to be one of the critical obstacles to the range view pipeline, and none of the previous range view detectors shows any positive results. This impedes the practical application of the range view pipeline that has many unique merits (e.g., avoiding voxelization/sparse convolutions/being lightweight and etc.). Moreover, our work opens up many new possibilities for this promising direction. For example, our method can enable the range view detectors to predict the velocity of the target objects, which is previously infeasible. On the nuScenes val set, our best model attains an averaged velocity error (AVE) = $0.301$, being significantly better than the single-frame model AVE = $1.08$.

\textbf{Q. The ablation study of Table 3.}

All the items in Table 3 are with the multi-frame fusion, and the difference is the way to group the channels. 
All the inference time is measured with batch size 1 on a 3090Ti GPU. We do not use a sliding temporal window, and our model only simply forwards the range view images through the network. Additionally, the time differences in Table 3 (except the 1st row) are small ($\leq$1ms), which might be dominated by the underlying implementations instead of the network architectures. Thus, we hope you can pay more attention to the model's performance here, and what we choose is the model with the best mAP and NDS.

\textbf{Q. Comparisons with the nuScenes leaderboard results.}

We 
 want to note that the methods on the leaderboard often make use of model ensembles, test-time augmentation and other tricks (e.g., PointPainting) to attain good performance. Our method has no bells and whistles and can be easily deployed in practice. Moreover, the results on the leaderboard are not peer-reviewed and thus we mainly compare the results from published papers.

\textbf{Q. How to handle the pixels each having multiple points?}

Yes, a single pixel on the range image can have multiple 3D points due to the multi-round projection, which might belong to different objects. In this work, we only use the points in the \emph{first-round projection} to compute the target assignments. We did attempt to use the points from all rounds, which requires an additional output branch to predict which round is active for each pixel in inference. However, both achieved a similar performance, as shown in the following table. We conjecture that the similar performance is due to the fact that the points in the first-round projection are already enough to assign at least one positive sample to each target object.

\textbf{Q. How to find the points corresponding to a given range view pixel?}

Although each location $(x, y)$ on the FPN feature maps with downsampling ratio $s > 1$ can be mapped to multiple points, we prescribe it is mapped to one exact pixel $(xs, ys)$ on the range image. One pixel on the range image corresponds to one 3D point (in the first-round projection) and thus there is no ambiguity that points with different depths are simultaneously assigned to one pixel. Then, we obtain the corresponding 3D point at this location and check whether the 3D point is inside any ground-truth box. If any, the ground-truth box is assigned as the target of this pixel. Otherwise, the pixel is background. If a point falls in multiple ground-truth boxes (this case is very rare in 3D object detection), the one with the minimum area is chosen.

\begin{table}[h!]
\centering 
\footnotesize 
 \setlength{\tabcolsep}{2pt}
\caption{
\textbf{Whether to use all rounds of projection to compute target assignments.}}
  \label{table:all_round}
  \begin{tabular}{cccccccccccccc}
     & mAP(\%) & NDS%
    (\%) & Car & Truck & Bus & Trailer & C.V. & Ped. &  Motor & Bicycle & T.C. & Barrier \\
    \hline
    all rounds & 58.18 & 64.27 & 83.4 & 53.4 & 67.2 & 33.3 & 19.6 & 83.7 & 60.6 & \textbf{37.8} & \textbf{73.0} & \textbf{69.7} \\
    first round only & \textbf{58.49} & \textbf{64.64} & \textbf{83.5} & \textbf{54.6} & \textbf{67.4} & \textbf{34.9} & \textbf{19.7} & \textbf{84.2} & \textbf{62.2} & 36.3 & 72.4 & \textbf{69.7} \\
  \end{tabular}
\end{table}



{
\small
\bibliographystyle{alpha}

\bibliography{bibtex}
}

\end{document}